\documentclass[letterpaper, 10 pt, conference]{ieeeconf}

\usepackage{times}
\usepackage{multicol}
\usepackage[bookmarks=true]{hyperref}

\usepackage{amsmath} 
\usepackage{amsfonts}
\usepackage{amssymb} 
\usepackage{mathtools} 
\usepackage{graphicx}
\usepackage{subcaption} 
\usepackage{float} 
\usepackage{bm} 

\usepackage{enumerate} 
\usepackage[top=54pt, left=54pt, right=54pt, bottom=54pt]{geometry} 
\usepackage[dvipsnames]{xcolor}
\usepackage{tikz}
\usetikzlibrary{bayesnet}
\usepackage{float}
\usepackage{lipsum}
\usepackage{cite}

\DeclareMathOperator*{\argmax}{argmax}

\usepackage[skip=3pt,font=small,labelfont=bf]{caption} 
\newcommand{\shrinka}{\def\baselinestretch{0.993}\large\normalsize}
\title{\LARGE \bf Active Learning of Probabilistic Movement
  Primitives }

\IEEEoverridecommandlockouts 
\author{Adam Conkey$^1$ and Tucker Hermans$^{1,2}$%
\thanks{$^{1}$School of Computing; Robotics Center; University of Utah, USA. $^{2}$NVIDIA, USA. \emph{Email: adam.conkey@utah.edu, thermans@cs.utah.edu}}}

\begin{document}
\shrinka
\maketitle
\shrinka

\begin{abstract}

  A Probabilistic Movement Primitive (ProMP) defines a distribution over
  trajectories with an associated feedback policy. ProMPs are typically
  initialized from human demonstrations and achieve task generalization through
  probabilistic operations. However, there is currently no principled guidance
  in the literature to determine how many demonstrations a teacher should
  provide and what constitutes a ``good'' demonstration for promoting
  generalization. In this paper, we present an active learning approach to
  learning a library of ProMPs capable of task generalization over a given
  space. We utilize uncertainty sampling techniques to generate a task instance
  for which a teacher should provide a demonstration.  The provided
  demonstration is incorporated into an existing ProMP if possible, or a new
  ProMP is created from the demonstration if it is determined that it is too
  dissimilar from existing demonstrations. We provide a qualitative comparison
  between common active learning metrics; motivated by this comparison we
  present a novel uncertainty sampling approach named ``Greatest Mahalanobis
  Distance.'' We perform grasping experiments on a real KUKA robot and show our
  novel active learning measure achieves better task generalization with fewer
  demonstrations than a random sampling over the space.

\end{abstract}

\section{Introduction}
\label{sec:intro}

Learning from demonstration~\cite{atkeson1997robot, billard2008robot,
argall2009survey} offers a promising approach for robot users untrained in
programming to command robots to perform common manipulation tasks. By teaching
the robot through demonstration, the user can provide manipulation expertise
without needing to be an expert in robotics.  Probabilistic Movement Primitives
(ProMPs) provide a useful policy representation for generating adaptable robot
motion learned from demonstration~\cite{paraschos2018using}. A ProMP encodes a
distribution over trajectories and is typically initialized with several
demonstrations from a human teacher. Task generalization to new goals and
contexts is primarily achieved by conditioning the trajectory distribution on
desired trajectory waypoints. This generalization mechanism has been
successfully applied in a variety of applications including grasping objects
while avoiding obstacles~\cite{paraschos2017prioritization}, relocating objects
of unknown weight~\cite{paraschos2018probabilistic}, collaborative assembly
tasks~\cite{ewerton2015learning}, and robot table
tennis~\cite{gomez-gonzalez2016using}.

However, ProMPs require an indeterminate number of demonstrations to confidently
generalize over the desired task space and appropriately estimate the associated
task covariance~\cite{maeda2017active}. Many real-world tasks require hundreds
of demonstrations to fully estimate the demonstration
covariance~\cite{colome2017demonstration}. If too few demonstrations are
provided, numerical issues arise in the form of singular covariance matrices,
and it is common to use a non-informative prior for the covariance in order to
sidestep this issue~\cite{paraschos2018using, wilbers2017context,
havoutis2017supervisory}. However, the generalization capability of a ProMP can
be compromised when using parameters that do not adequately estimate the true
distribution associated with a task~\cite{paraschos2018using,
havoutis2017supervisory}. The current alternatives available to the teacher are
to either expend undue effort to exhaustively demonstrate a task to the robot,
or to attempt to capture, in only a small number of demonstrations, the task
variation necessary to achieve the desired generalization. There is a need for a
third option that guides the teacher to provide only those demonstrations
necessary to ensure the desired task generalization is achievable.

\begin{figure}
  \includegraphics[width=0.48\textwidth]{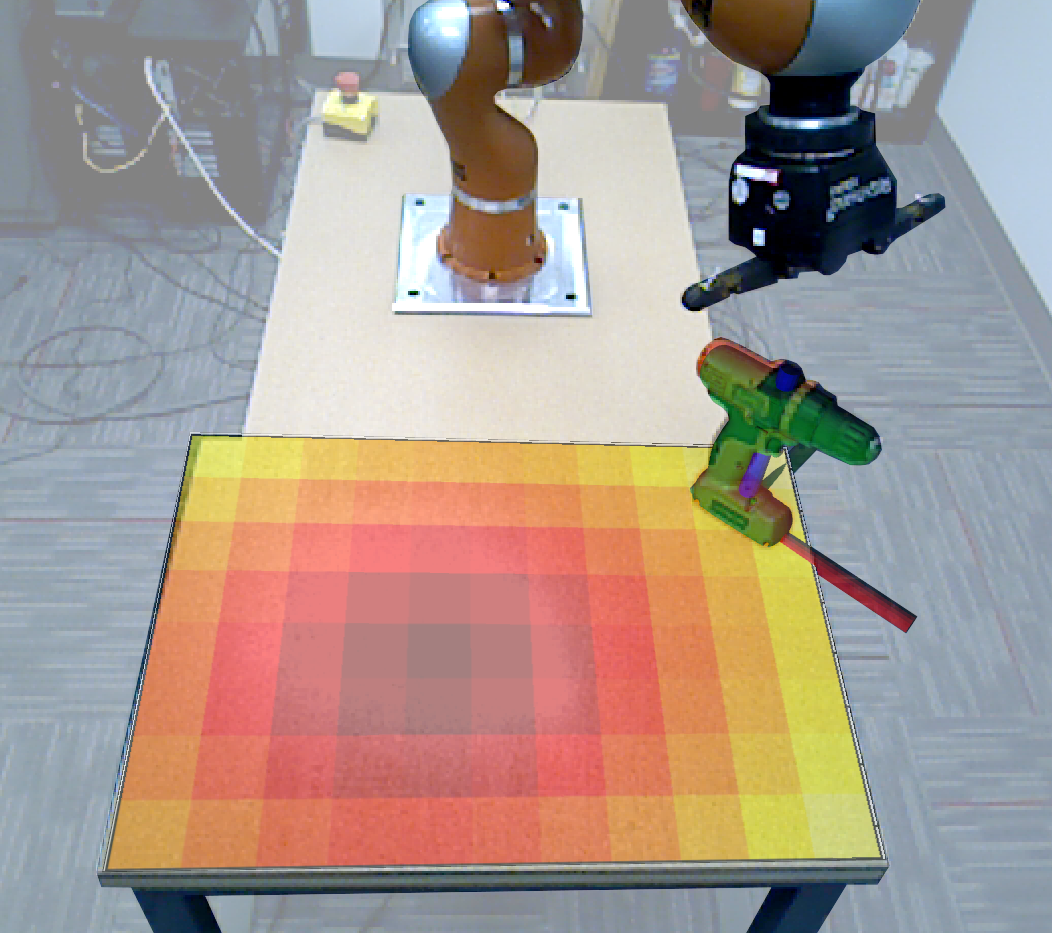}
  \caption{Illustration of our experimental task and setup. The robot attempts
  to pick up an object placed anywhere on the table surface. The system
  autonomously generates the task instance it is least likely to generalize to
  according to an active learning metric. The teacher then provides a
  demonstration for how to pick up the object from the designated location.}
  \label{fig:cover}
\end{figure}  

In this paper, we present a novel active learning procedure for learning a
library of ProMPs from demonstration that is capable of task generalization over
a desired region. We frame this as an active learning problem by conceiving of
each ProMP in the library as its own class, with the guiding intuition that we
want to fully ``classify'' the space, i.e. achieve full ProMP coverage of the
space. We adopt an uncertainty sampling approach~\cite{settles2012active} that
enables the robot to generate the task sample for which it is least likely to
``predict'' correctly, i.e. generalize to with a ProMP. By allowing the learner
to generate task samples to be ``labeled'', i.e. demonstrated by the teacher, we
remove the burden of the teacher to decide which demonstration to provide
next. Additionally, by informing task selection with uncertainty measures, we
reduce the total number of demonstrations necessary to achieve a task than if
demonstrations are given in an ad hoc manner.

We provide a qualitative comparison of different uncertainty sampling measures
commonly used for active learning in supervised learning settings: Least
Confident, Maximum Entropy, and Minimum Margin. We show that these measures are
not suitable for promoting task generalization with ProMPs. We propose a new
measure we call Greatest Mahalanobis Distance that effectively generates task
instances that are not in close proximity to any existing ProMP distribution. We
demonstrate with grasping experiments on a real KUKA robot that our method
affords task generalization with fewer demonstrations more effectively than
randomly sampling over the space.

To briefly summarize our contributions, in this paper we:
\begin{enumerate}
  \item Formalize an active learning approach for learning manipulation tasks
  from demonstration using Probabilistic Movement Primitives.
  \item Provide qualitative comparisons of the three most common uncertainty
  sampling techniques: Least Confident, Minimum Margin, and Maximum Entropy.
  \item Present a novel uncertainty sampling function suitable for building a
  mixture of ProMPs capable of generalizing a manipulation task within a given
  region.
  \item Leverage the probabilistic information encoded in ProMP policies to
  automatically determine which ProMP in the mixture a new demonstration should
  be incorporated into, and which ProMP to execute for a new task instance.
\end{enumerate}

We structure the remainder of the paper as follows. We first review related work
in active learning from demonstration in Section~\ref{sec:related_work}. We then
provide a brief technical overview of ProMPs in Section
\ref{sec:methods:background} and define our novel approach to learning a library
of ProMPs through active learning in
Sections~\ref{sec:methods:mixture}--\ref{sec:methods:promp_context}. We present
an overview of our experimental setup in Section~\ref{sec:experiment_setup} and
describe the corresponding results in Section~\ref{sec:experiment_results} for
grasping experiments performed on a physical KUKA LBR4+ robot. We conclude in
Section~\ref{sec:conclusion} with some final remarks and directions for future
work.


\section{Related Work}
\label{sec:related_work}

Active learning, where a learner actively poses queries to a teacher for input
to reduce sample complexity, has been widely applied in supervised learning
settings \cite{settles2012active}. Our approach is most suitably situated in the
literature on \textit{active learning from demonstration} \cite{maeda2017active,
silver2012active, grollman2007dogged, kroemer2010combining,chen2018active}, also
referred to as \textit{active imitation learning}~\cite{shon2007active,
judah2014active}, in which the learner generates task instances for which the
teacher may provide a demonstration. Active learning from demonstration has been
applied to autonomous navigation \cite{silver2012active}, object seeking with a
quadruped robot \cite{grollman2007dogged}, grasping objects
\cite{kroemer2010combining}, reaching to task space positions with a manipulator
\cite{maeda2017active}, and generating smooth robot motion from a latent-space
encoding~\cite{chen2018active}. Also included in this area are approaches where
the learner does not request full task demonstrations, but instead asks for the
action to take in the particular state that it is in \cite{shon2007active,
judah2014active, chernova2009interactive}. These approaches, however, are only
applicable to finite action spaces where actions are easily specified by a
teacher.

The approaches most closely related to ours are those using active learning to
learn Dynamic Movement Primitives (DMPs) for grasping
\cite{kroemer2010combining} and reaching tasks \cite{maeda2017active}. In
\cite{kroemer2010combining}, a hybrid system is presented such that a high-level
active learner generates grasp configurations based on a variant of Upper
Confidence Bound (UCB) policies \cite{sutton2018reinforcement}, and a low-level
reactive DMP controller executes the grasp motion based on a task
demonstration. In \cite{maeda2017active}, the robot incrementally learns DMPs
for reaching pre-defined positions in its workspace. A Gaussian process is used
for sampling trajectories with an associated variance. If a function of the
variance is below an uncertainty threshold, then an existing DMP is used with
the goal appropriately adapted. Otherwise, the human user is asked for a new
demonstration to reach the new goal position. By utilizing ProMPs in our method
instead of DMPs, we are able to achieve greater generalization capabilities
\cite{paraschos2018using} while leveraging the probabilistic information already
encoded in the policy representation to compute confidence
measures. Additionally, we are able to provide a probabilistic measure of the
robot's ability to generalize a task in a given region, as opposed to
\cite{maeda2017active} which can only say for a given instance whether or not
the robot is confident it can execute the motion. Our approach therefore has the
advantage that the robot, after learning a task, can be deployed with an
associated uncertainty estimate that it will succeed on any task instance it is
given to perform.

Also relevant to our approach is work in the area of active learning for
parameterized skills \cite{daSilva2014active}. In \cite{daSilva2014active} the
agent selects tasks to practice in a reinforcement learning setting with the
objective of optimizing for expected improvement in skill performance. Task
competency is measured over a recursively split goal space in
\cite{baranes2013active} for an intrinsically-motivated agent. Active Contextual
Policy Search \cite{fabisch2014active} considers a learner that generates task
contexts to condition a high-level policy on, such that a lower-level policy can
be optimized to maximize an intrinsic reward function. These works are each
applied in reinforcement learning settings and are agnostic to any particular
policy representation. Our approach, on the other hand, makes use of human
demonstrations and, by committing to a particular policy representation
(ProMPs), we are able to compute task competency in a unified manner utilizing
information from the policy representation itself.


\section{Methods}
\label{sec:methods}

We first provide a brief background on ProMPs in
Section~\ref{sec:methods:background} to introduce the concepts relevant to our
contributions. We describe our approach to learning a mixture of ProMPs in
Section~\ref{sec:methods:mixture}. In Section~\ref{sec:methods:active_learning}
we present our novel approach for active learning of ProMPs and discuss the
methods we compare. Finally, we provide details in
Section~\ref{sec:methods:promp_context} on how to use our active learning method
for the concrete task we consider in our experiments: reaching to grasp an
object.

\subsection{Background}
\label{sec:methods:background}
We utilize a formulation of ProMPs that closely parallels that of
\cite{paraschos2018using}. The ProMP trajectory distribution has the general
form
\begin{equation}
\label{eqn:prob_trajectory}
  p(\bm{\tau} \mid \bm{w}, \bm{\Sigma}_{\bm{y}}) = \prod_t p(\bm{y}_t \mid \bm{\Psi}_t \bm{w}, \bm{\Sigma}_{\bm{y}}) \\
\end{equation}
where $\bm{\tau} = [\bm{y}_0, \dots, \bm{y}_T]$ is a trajectory of the state
$\bm{y}_t \in \mathcal{S}$ for state space $\mathcal{S}\subseteq\mathbb{R}^d$,
$\bm{\Psi}_t \in \mathbb{R}^{d \times dn}$ is a block-diagonal matrix of $n$
basis functions for each dimension of the state, $\bm{w} \in \mathbb{R}^{dn}$ is
a weight vector, and $\bm{\Sigma}_{\bm{y}}$ is the observation noise. We assume,
as in \cite{paraschos2018using}, that the time-dependent distributions are
Gaussian, i.e.
$p(\bm{y}_t \mid \bm{\Psi}_t \bm{w}, \bm{\Sigma}_{\bm{y}}) =
\mathcal{N}(\bm{y}_t \mid \bm{\Psi}_t \bm{w}, \bm{\Sigma}_{\bm{y}})$. This
results in $p(\bm{\tau} \mid \bm{w}, \bm{\Sigma}_{\bm{y}})$ being Gaussian as
well since it is a product of Gaussian distributions.

We parameterize the distribution with
$\bm{\theta} = \{\bm{\mu}_{\bm{w}}, \bm{\Sigma}_{\bm{w}}\}$ and marginalize out
the weights such that
\begin{align}
  p(\bm{\tau} \mid \bm{\theta}, \bm{\Sigma}_{\bm{y}}) &= \int p(\bm{\tau} \mid \bm{w}, \bm{\Sigma}_{\bm{y}}) p(\bm{w} \mid \bm{\theta}) d\bm{w}
\end{align}
Task generalization is achieved by conditioning $p(\bm{w} \mid \bm{\theta})$ on
a desired trajectory waypoint $\bm{y}_t^*$ with covariance
$\bm{\Sigma}_{\bm{y}_t}^*$. The updated parameters
$\bm{\theta}^* = \{\bm{\mu}_{\bm{w}}^*, \bm{\Sigma}_{\bm{w}}^*\}$ are computed
by
\begin{align}
\bm{\mu}_{\bm{w}}^* &= \bm{\mu}_{\bm{w}} + \bm{\Sigma}_{\bm{w}} \bm{\Psi}_t\left(\bm{\Sigma}_{\bm{y}_t}^*
+ \bm{\Psi}_t^T\bm{\Sigma}_{\bm{w}}\bm{\Psi}_t\right)^{-1}(\bm{y}_t^*
- \bm{\Psi}_t^T\bm{\mu}_{\bm{w}}) \label{eqn:condition_mu} \\
\bm{\Sigma}_{\bm{w}}^* &= \bm{\Sigma}_{\bm{w}} - \bm{\Sigma}_{\bm{w}} \bm{\Psi}_t\left(\bm{\Sigma}_{\bm{y}_t}^*
+ \bm{\Psi}_t^T\bm{\Sigma}_{\bm{w}}\bm{\Psi}_t\right)^{-1}\bm{\Psi}_t^T\bm{\Sigma}_{\bm{w}} \label{eqn:condition_sigma}
\end{align}
This closed-form update is possible because we assume, as in
\cite{paraschos2018using}, that $p(\bm{w} \mid \bm{\theta})$ is Gaussian.

\subsection{Learning a Mixture of ProMPs from Demonstration}
\label{sec:methods:mixture}

We employ a mixture of multiple ProMPs parameterized as
$\mathcal{M} = \{(\bm{\theta}_1,\pi_1), \dots, (\bm{\theta}_J,\pi_J)\}$ where
$\bm{\theta}_j = \{\bm{\mu}_{\bm{w}}^j, \bm{\Sigma}_{\bm{w}}^j\}$, since it is
known that a single ProMP is not sufficient to properly characterize a given
space~\cite{ewerton2015learning}. We formalize the mixture as
\begin{equation}
\label{eqn:promp_mixture}
p(\bm{\tau} \mid \mathcal{M}) = \sum_{j=1}^J \pi_j \mathcal{N}(\bm{\tau} \mid \bm{\Psi} \bm{\mu}_{\bm{w}}^j, \bm{\Psi}^T\bm{\Sigma}_{\bm{w}}^j\bm{\Psi}+\bm{\Sigma}_{\bm{y}})
\end{equation}
where $\pi_j \in [0,1]$ are mixture coefficients and $\bm{\mu}_{\bm{w}}^j$,
$\bm{\Sigma}_{\bm{w}}^j$ are the mean and covariance associated with the
$j^{th}$ ProMP.

The mixture $\mathcal{M}$ is learned incrementally over time as new
demonstrations are acquired. In order to incorporate a new demonstration, we
first learn a weight vector $\bm{w}$ from the demonstration using Ridge
regression as in~\cite{paraschos2018using}. For the first demonstration
received, a new ProMP is created with mean $\bm{\mu}_{\bm{w}}^1 = \bm{w}$ and
covariance $\bm{\Sigma}_{\bm{w}}^1 = \gamma \mathbf{I}$, where $\mathbf{I}$ is
the identity matrix and $\gamma \in \mathbb{R}^+$ is a scaling parameter. This
serves as a non-informative prior for the covariance
\cite{paraschos2018using}. For subsequent demonstrations, we must determine
which ProMP in the mixture the new demonstration should be incorporated into. In
contrast to previous work~\cite{koert2018incremental} that learns a separate
model for a gating function to the mixture components, we directly utilize the
probabilistic information encoded in the learned ProMPs to determine which ProMP
a new demonstration should be incorporated into. We use the Mahalanobis
distance~\cite{mahalanobis1936generalized} as a measure of disparity between
$\bm{w}$ and each ProMP distribution $\bm{\theta}_j$ given by
\begin{equation}
\label{eqn:mahalanobis}
d(\bm{w}, \bm{\theta}_j) = \sqrt{(\bm{w} - \bm{\mu}_{\bm{w}}^j)^T (\bm{\Sigma}_{\bm{w}}^j)^{-1}(\bm{w} - \bm{\mu}_{\bm{w}}^j)}
\end{equation}

A demonstration is incorporated into the $j^{th}$ ProMP if the Mahalanobis
distance between the learned weight vector and the ProMP distribution falls
below a disparity threshold $\delta \in \mathbb{R}^+$, i.e. if
$d(\bm{w}, \bm{\theta}_j) \leq \delta$. Instead of choosing a fixed threshold,
we compute a robust measure of a disparity threshold for each ProMP utilizing
the ProMP generative model. For each ProMP, we create a set of weight vector
samples $\mathcal{W}_j$ and compute the value of Equation
(\ref{eqn:mahalanobis}) for each $\bm{w}_i \in \mathcal{W}_j$. We use Median
Absolute Deviation outlier filtering~\cite{leys2013detecting} to compute the
threshold
\begin{equation}
\label{eqn:mad_filter}
\delta_j = \max\left\{d(\bm{w}_i, \bm{\theta}_j) : \frac{d(\bm{w}_i, \bm{\theta}_j) - M_j}{MAD_j} < \beta\right\}
\end{equation}
where $M_j$ is the median Mahalanobis distance of the sample weights
$\bm{w}_i \in \mathcal{W}_j$ to the ProMP distribution $\bm{\theta}_j$ and
$MAD_j$ is the Median Absolute Deviation computed by
$MAD_j = \text{med}\left(|d(\bm{w}_i, \bm{\theta}_j) - M_j|\right)$. The
parameter $\beta$ is an easily tuned parameter that governs how many outliers
are discarded and has standard associated values ranging from approximately 3
(few outliers discarded) to 2 (many outliers discarded)
\cite{leys2013detecting}.

Once it is determined that $d(\bm{w}, \bm{\theta}_j) \leq \delta_j$, the new
demonstration is incorporated into the $j^{th}$ ProMP by updating the ProMP's
distribution parameters as
\begin{align}
  \bm{\mu_w}^j &= \frac{1}{N} \sum_{i=1}^N \bm{w}_i \\
  \bm{\Sigma_w}^j &= \lambda \bm{\Sigma}_0 + \frac{(1-\lambda)}{N} \sum_{i=1}^N (\bm{w}_i - \bm{\mu_w}^j)(\bm{w}_i - \bm{\mu_w}^j)^T
\end{align}
The mean $\bm{\mu_w}^j$ is computed as the Maximum Likelihood Estimate (MLE)
where $N$ is the number of samples the ProMP is learned from, including the
newly acquired sample. $\bm{\Sigma_w}^j$ is updated as the Maximum A Posteriori
(MAP) estimate under an Inverse Wishart Prior, which amounts to a convex
combination of a positive semi-definite prior $\bm{\Sigma}_0$ and the MLE of the
sample covariance~\cite{murphy2012machine}. We adopt the method of
\cite{wilbers2017context} and set $\bm{\Sigma}_0$ to be the estimate of
$\bm{\Sigma_w}^j$ from the previous learning iteration. This ensures that
$\bm{\Sigma_w}^j$ is always full rank (due to the initial diagonal prior) and
that the parameter estimate is not unduly influenced by a new sample. We found
this to be important in our experiments since, in general, the number of
demonstrations each ProMP is learned from is considerably smaller than the
dimensionality of the weight space; using an ill-conditioned matrix in the
probability computations can result in non-informative values.

If it happens that $d(\bm{w}, \bm{\theta}_j) > \delta_j$ for every ProMP in the
mixture, then we create a new ProMP with an uninformative prior as described
previously. When initializing all ProMPs we set the initial
$\bm{\Sigma_{0}}=\sigma \bm{I}$ for a small value of $\sigma \in \mathbb{R}^+$.

\subsection{Active Learning of ProMPs}
\label{sec:methods:active_learning}
The active learner's objective is to learn a mixture of ProMPs $\mathcal{M}$
that achieves task generalization over some region of its environment. We
formalize this region by defining a continuous context space $\mathcal{C}$ that
specifies the task to be performed in terms of context variables
\cite{kober2011reinforcement} (e.g. the pose of an object to be grasped). We
assume there is a subset $\mathcal{C}_d \subseteq \mathcal{C}$ over which task
generalization is desired.

We estimate the achievable feasible region by the coverage achieved by the
mixture of ProMPs at the timestep relevant for the task context
$\bm{\eta}$. Because the context variable is not, in general, a direct subset of
the ProMP state, we allow for a mapping $g:\mathcal{C} \rightarrow \mathcal{S}$
between the context space $\mathcal{C}$ and the ProMP state space
$\mathcal{S}$. We define one such mapping below in
Section~\ref{sec:methods:promp_context} suitable for our experimental grasping
task.

We formalize our active learning problem by conceiving of each ProMP in the
mixture to be its own class. We then employ active learning through uncertainty
sampling~\cite{settles2012active}, in which the learner generates a new task
instance for which the teacher can provide a demonstration governed by
\begin{equation}
\bm{\eta}^* = \argmax_{\bm{\eta} \in \mathcal{C}_d} U(\bm{\eta})
\end{equation}
where $\bm{\eta} \in \mathcal{C}_d$ is a context variable sufficient to describe
the task and $U(\bm{\eta})$ is an uncertainty sampling function that measures
the uncertainty the learner has about characterizing a given task instance as
being a member of one of the available classes. We qualitatively compare the
three most common uncertainty sampling measures \cite{settles2012active}:

\noindent\textbf{Least Confident:}
\begin{equation}
\label{eqn:least_confident}
U_{\tiny{lc}}(\bm{\eta}) = \argmax_{\bm{\eta} \in \mathcal{C}_d} \left[1 - p(z_1 \mid \bm{\eta}) \right]
\end{equation}

\noindent\textbf{Minimum Margin:}
\begin{equation}
\label{eqn:min_margin}
U_{mm}(\bm{\eta}) = \argmax_{\bm{\eta} \in \mathcal{C}_d}\left[p(z_2 \mid \bm{\eta}) - p(z_1 \mid \bm{\eta})\right]
\end{equation}

\noindent\textbf{Maximum Entropy:}
\begin{equation}
\label{eqn:max_entropy}
U_{me}(\bm{\eta}) = \argmax_{\bm{\eta} \in \mathcal{C}_d} -\sum_{z = 1}^{J} p(z \mid \bm{\eta}) \log p(z \mid \bm{\eta})
\end{equation}

In Equations \ref{eqn:least_confident}--\ref{eqn:max_entropy},
$p(z \mid \bm{\eta})$ indicates the probability of a class label $z$ being
attributed to instance $\bm{\eta}$, where the class label corresponds to any one
of the $J$ ProMPs. In Equations \ref{eqn:least_confident} and
\ref{eqn:min_margin}, $z_1 = \argmax_z p(z \mid \bm{\eta})$ is the most likely
label for instance $\bm{\eta}$ while $z_2$ in Eq.~\ref{eqn:min_margin} is the
second most likely label. Intuitively, the Least Confident measure
(Eq.~\ref{eqn:least_confident}) selects the task instance \(\bm{\eta}^*\) whose
highest probability over all labels $z \in \mathcal{Z}$ is lowest compared to
all other instances $\bm{\eta} \in \mathcal{C}_d$. The Minimum Margin measure
(Eq.~\ref{eqn:min_margin}) chooses the instance with the greatest ambiguity
between its two most likely classifications. The Maximum Entropy measure
(Eq.~\ref{eqn:max_entropy}) identifies the instance with the highest label
uncertainty over all classes.

We define an additional, novel uncertainty sampling function:

\noindent\textbf{Greatest Mahalanobis Distance}
\begin{equation}
    \label{eqn:greatest_mahalanobis}
    U_{gm}(\bm{\eta}) = \argmax_{\bm{\eta} \in \mathcal{C}_d} \min_{j} d\left(\bm{\eta}, \bm{\theta}_{\bm{\eta}}^j\right)
\end{equation}
where $d(\cdot)$ is the Mahalanobis distance defined in
Equation~\ref{eqn:mahalanobis}, $j$ indexes over ProMPs, and
$\bm{\theta}_{\bm{\eta}}^j = \{\bm{\mu}_{\bm{\eta}}^j,
\bm{\Sigma}_{\bm{\eta}}^j\}$ defines a distribution over the context variable
achieved by mapping the $j^{th}$ ProMP distribution parameters to the context
space. We provide details for the specific mapping we utilize in this paper
below in Section~\ref{sec:methods:promp_context}. Our Greatest Mahalanobis
Distance approach is similar to Least Confident, but instead of choosing the
instance with the lowest probability over classes (ProMPs), it selects the
instance whose closest ProMP distribution is the farthest
away.

We found that in practice, the Mahalanobis distance is less susceptible to
computational issues than the probability values computed for the other
uncertainty sampling functions. The density function for a Gaussian distribution
requires dividing by the determinant of the covariance matrix, which is
equivalent to dividing by the product of the eigenvalues of the covariance
matrix. This value can be extremely small when the covariance is estimated from
a small sample set, causing the computation to become unstable. We show in our
experiments in Section~\ref{sec:experiment_results} that the Greatest
Mahalanobis Distance objective encourages the learner to select instances far
away from instances it has already received demonstrations for, while the other
uncertainty sampling functions tend to ``compete'' along the boundaries of the
regions covered by adjacent ProMPs.

Given the new task instance $\bm{\eta}^*$ generated by the uncertainty sampling
optimization, the teacher provides a demonstration. The demonstration is then
incorporated into the mixture of ProMPs as described in
Section~\ref{sec:methods:mixture}. The procedure iterates until a stopping
criteria is met, e.g. the task success rate over a validation set reaches an
acceptable percentage.

\subsection{Example ProMP Context}
\label{sec:methods:promp_context}

To be concrete in our formulation, we present a context mapping for the task of
grasping an object placed arbitrarily on a surface. We use this mapping in our
experiments presented later in Section~\ref{sec:experiment_results}. The task
requires the robot to pick up an object located arbitrarily on a table
surface. The ProMP state consists of the end-effector pose with respect to the
robot's base frame $^{0}T_{ee}$ (e.g. from forward kinematics of the joint
state), while the context space is the pose of the object with respect to the
base frame $^{0}T_{obj}$ (e.g. from an object tracker using an RGB-D
camera~\cite{wuthrich2013probabilistic}). Once a desired end-effector pose in
the object frame $^{obj}T_{ee}$ is known, the mapping
$g:\mathcal{C}\rightarrow\mathcal{S}$ from context space to state space, as
described in Section~\ref{sec:methods:active_learning}, is achieved by a simple
coordinate frame transformation:
 \begin{equation}
     g\left(^{0}T_{obj}\right) = {^{0}T_{obj}} \cdot  {^{obj}T_{ee}} = {^{0}T_{ee}}
 \end{equation}

\begin{figure}
  \centering
  \includegraphics[width=0.37\textwidth]{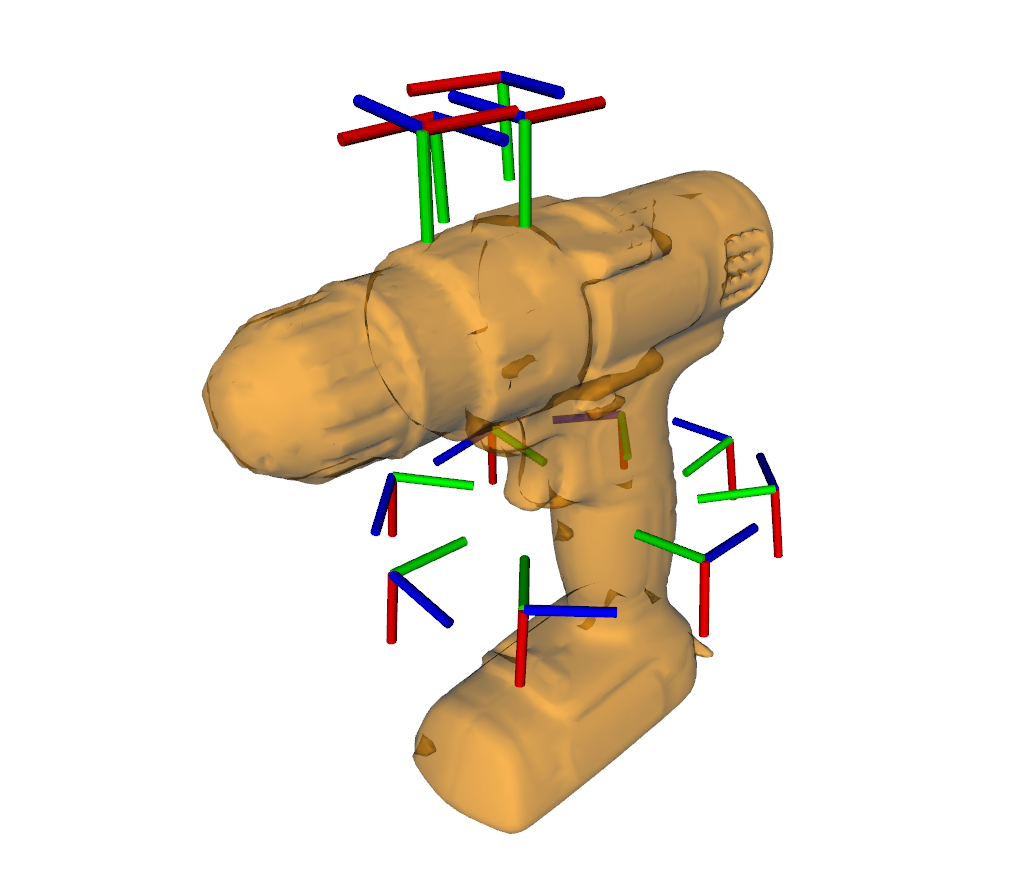}
  \caption{Mean poses of the GMM over end-effector poses in the object frame
  used for conditioning the ProMP distribution.}
  \label{fig:ee_gmm}
\end{figure}

 The pose $^{obj}T_{ee}$ could be specified manually or from the output of a
 grasp planner; however, we instead employ a Gaussian Mixture Model (GMM) over
 successful end-effector poses in the object frame. The GMM is defined by
\begin{equation}
\label{eqn:ee_gmm}
p(\bm{y}_t) = \sum_{r=1}^R \beta_r \mathcal{N}(\bm{y}_t \mid \bm{\mu}_{\bm{y}_t}^r, \bm{\Sigma}_{\bm{y}_t}^r)
\end{equation}
where $\beta_r \in [0,1]$ are the mixture coefficients and
$\bm{\mu}_{\bm{y}_t}^r, \bm{\Sigma}_{\bm{y}_t}^r$ are the mean and covariance of
the end-effector pose in the object frame for the $r^{th}$ component. A
visualization of the mean components learned from the demonstrations given in
our experiments can be seen in Figure~\ref{fig:ee_gmm}. Using the known pose of
the object in the base frame, we transform each $\bm{\mu}_{\bm{y}_t}^r$,
$\bm{\Sigma}_{\bm{y}_t}^r$ to get $\bm{\tilde \mu}_{\bm{y}_t}^r$,
$\bm{\tilde \Sigma}_{\bm{y}_t}^r$, which are the mean and covariance of the
end-effector with respect to the base frame.

We leverage these parameters as the condition points for the ProMP, i.e. we set
$\bm{y}_t^* = \bm{\tilde \mu}_{\bm{y}_t}^r$ and
$\bm{\Sigma}_{\bm{y}_t}^* = \bm{\tilde \Sigma}_{\bm{y}_t}^r$ in Equations
\ref{eqn:condition_mu} and \ref{eqn:condition_sigma}. We then compute the
probability of a particular task being achievable by the ProMP mixture as
\begin{equation}
\label{eqn:p_feasible}
p(\bm{\eta} \mid z=j) = \sum_{r=1}^R \beta_r \mathcal{N}(\bm{\tilde y}_t \mid \bm{\Psi}_t \bm{\tilde{\mu}}_{\bm{w}}^j, \bm{\Psi}_t^T \bm{\tilde{\Sigma}}_{\bm{w}}^j \bm{\Psi}_t+\bm{\Sigma}_y)
\end{equation}
where $z=j$ indicates the $j^{th}$ ProMP in the mixture; $\bm{\tilde y}$ is the
ProMP state generated from the transformation of context variable;
$\bm{\tilde{\mu}}_{\bm{w}}^j$ and $\bm{\tilde{\Sigma}}_{\bm{w}}^j$ are the
posterior distribution parameters in weight space computed from Equations
\ref{eqn:condition_mu} and \ref{eqn:condition_sigma}; and $\beta_r$ are the same
as in Equation~\ref{eqn:ee_gmm}.

We interpret Equation~\ref{eqn:p_feasible} as a measure of how capable the ProMP
is of achieving the task when conditioned on the task-relevant pose determined
by the context variable. There is little guidance in the literature for how to
set $\bm{\Sigma}_{\bm{y}_t}^*$ and it is typically taken to be a scaled identity
matrix~\cite{gomez-gonzalez2018adaptation}. We highlight this key advantage of
our choice to learn the GMM: we obtain meaningful values for both the mean and
the covariance for use in this conditioning operation.

We note that we are not able to directly compute the probabilities
$p(z \mid \bm{\eta})$ for the uncertainty sampling measures
(Eqs.~\ref{eqn:least_confident}--\ref{eqn:max_entropy}). Thus we use Bayes
theorem and Eq.~\ref{eqn:p_feasible} giving
\begin{align}
p(z \mid \bm{\eta}) &= \frac{p(\bm{\eta} \mid z)p(z)}{p(\bm{\eta})}
= \frac{p(\bm{\eta} \mid z)p(z)}{\sum_{z_i}p(\bm{\eta} \mid z_i)p(z_i)}
\end{align}
where $z_i$ ranges over all possible classes. We use a uniform, uninformative
prior for $p(z)$ to reflect our assumption that without further knowledge, any
ProMP in the mixture might potentially be used to execute a task. More
intelligent priors are worth exploring and we leave this for future work.


\section{Experimental Setup}
\label{sec:experiment_setup}

\begin{figure*}
  \centering
  \includegraphics[width=0.97\textwidth]{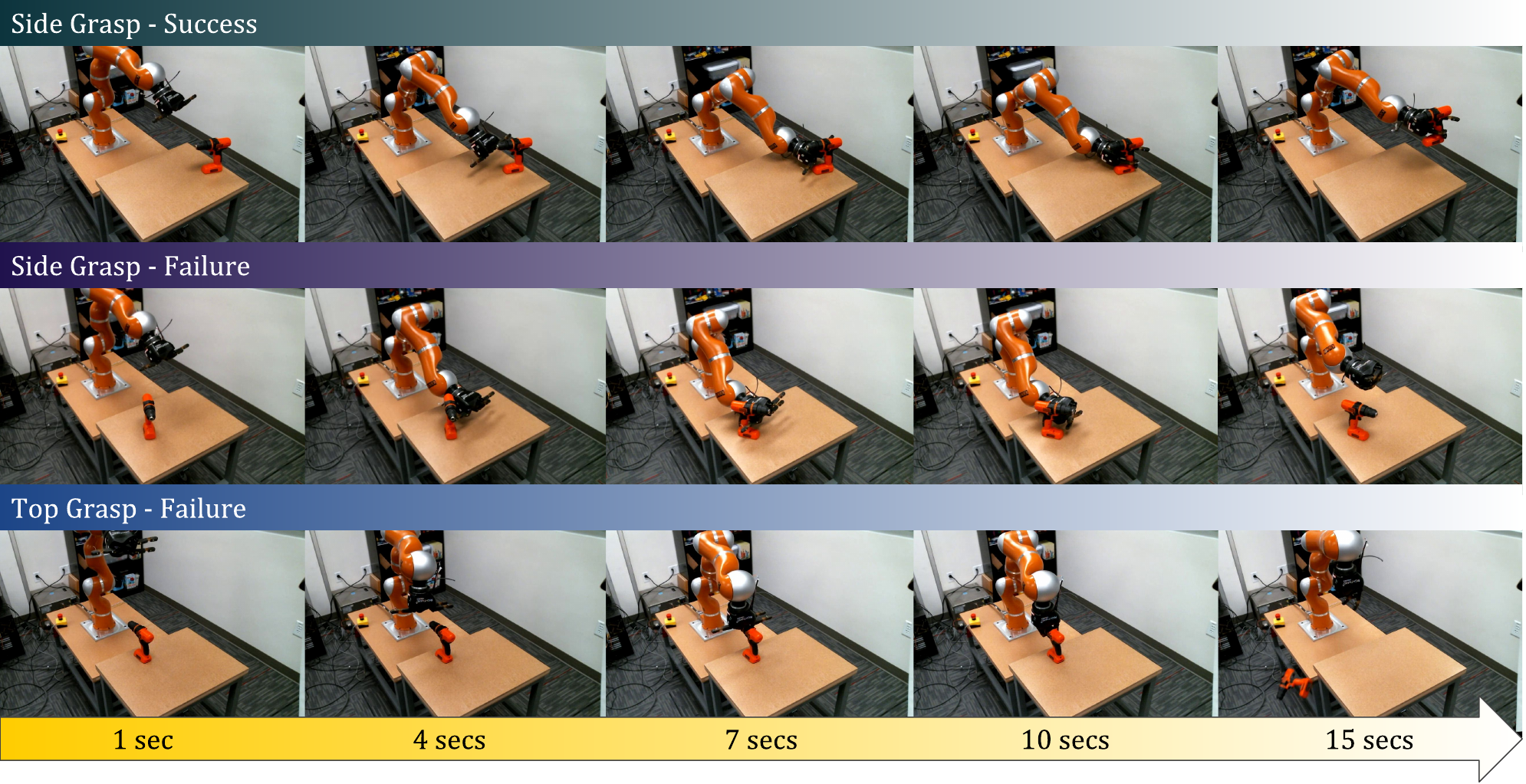}
  \caption{Characteristic execution outcomes of our experiments described in
  Section~\ref{sec:experiment_results:execution}. The first row is a successful
  side-grasp performed using our Greatest Mahalanobis Distance approach. The
  second row is a failed instance from the random-selection method that collides
  with the object and fails to pick it up. The final row is a failed instance
  from our method in which a more difficult top grasp was attempted, and the
  robot dropped the object during the lifting phase. Please see our attached
  video for more examples.}
  \label{fig:sequences}
\end{figure*}

We illustrate the qualitative differences of the active learning strategies
under consideration using a simple grasping task. The goal is for the robot to
be able to pick up a drill placed in an arbitrary planar pose on a table in the
robot's reachable workspace, as illustrated in Figures~\ref{fig:cover}
and~\ref{fig:sequences}. We chose this task because it affords an easily
discernible comparison of the different methods while providing a non-trivial
space to optimize over. In order to maintain consistency in the demonstrations
available to each comparison method, we discretized the sampling space into a
grid with planar positions in 5cm intervals and planar orientations in
increments of 45 degrees. The result is a total of approximately 700 possible
planar poses for selection. We provided one demonstration for each of these
samples through kinesthetic teaching of the robot in gravity compensation mode.

We provide a qualitative characterization of the three uncertainty sampling
methods discussed in Section \ref{sec:methods:active_learning}; namely, Least
Confident, Minimum Margin, and Maximum Entropy. We show that each of these
measures computed over the ProMP probabilities exhibits undesired behavior in
the context of active learning for ProMPs. We then provide a more rigorous
quantitative analysis comparing our proposed method of Greatest Mahalanobis
Distance to a random-selection strategy. We present results of executing the
grasping task using both methods and show that our method provides better task
generalization over the space with fewer demonstrations required from the
teacher.

We performed our experiments\footnote{Data is available at
\texttt{\url{http://bit.ly/al_promp_data}}.}\footnote{Code is available at
\texttt{\url{http://bit.ly/al_promp_code}}.}\footnote{Video is available at
\texttt{\url{https://youtu.be/na91UyidDvE}}.} on a KUKA LBR4+ robot arm equipped
with a ReFlex TakkTile hand \cite{odhner2014compliant,reflex_website}. Given the
Cartesian waypoints generated from a ProMP policy, we formulate a Sequential
Quadratic Program to obtain a joint trajectory by minimizing the L2 squared
error between the end-effector pose and the Cartesian
waypoints~\cite{sundaralingam2019relaxed}. We tracked the resulting trajectory
with a real-time Orocos~\cite{bruyninckx2001open} joint space PD controller
operated at 500Hz. Grasps were performed by assuming a canonical preshape and
closing the hand until contact was made (as detected by the
TakkTile~\cite{tenzer2014takktile} pressure sensors on the ReFlex fingers). We
then drove the motors a small additional amount to achieve a firm grasp,
following the control approach from~\cite{jentoft2014limits}. Once grasped, a
pre-defined lifting sequence was executed to lift the object approximately 20cm
above the table. A grasp is considered successful if the object is still in the
robot's grasp at the end of the lifting sequence.

Prior to executing any trajectory on the physical robot, we perform a kinematic
simulation of the robot with the environment model overlaid in rviz. We do not
execute any trajectory that is clearly dangerous in terms of colliding with the
environment at a non-trivial velocity.

We use the drill from the YCB dataset \cite{calli2015ycb} as the object to be
grasped by the robot, as shown in Figure~\ref{fig:cover}. We track the pose of
the object using the Bayesian object tracker described in
\cite{wuthrich2013probabilistic}. The pose is visualized in rviz and overlaid on
the camera feed coming from an ASUS X-tion Pro RGB-D camera. Selected task poses
for the object are also displayed in this way, and the human user utilizes the
displays to align the object pose with the generated task instance pose.


\begin{figure}
  \centering
  \begin{subfigure}[b]{0.25\textwidth}
    \includegraphics[width=\textwidth]{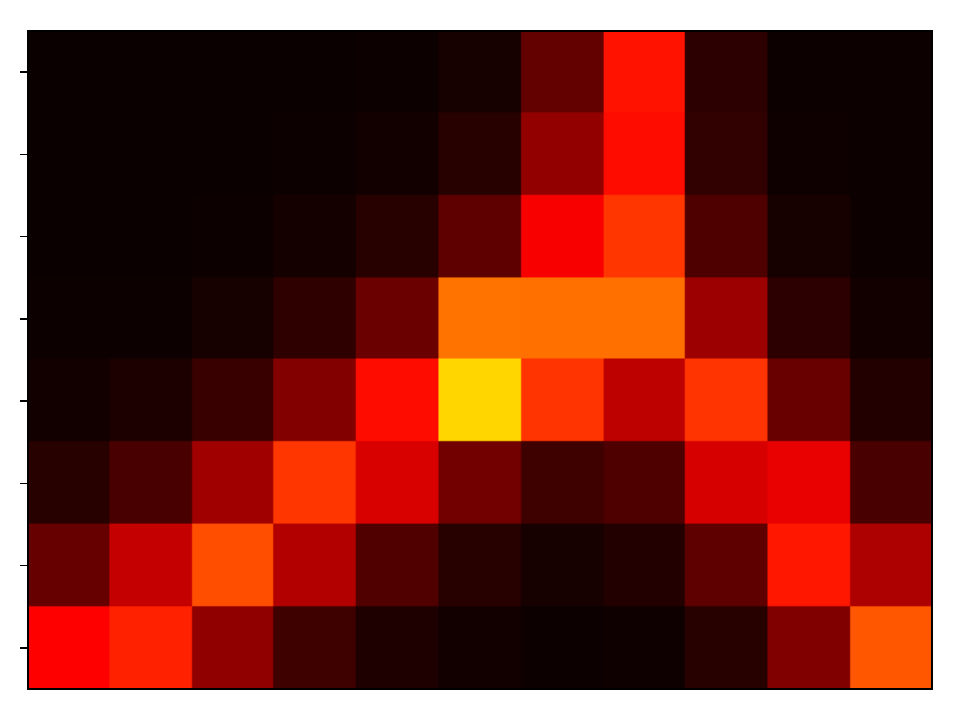}
    \caption{Multi-class ProMP}
    \label{fig:sampling:multi_class}
  \end{subfigure}
  \begin{subfigure}[b]{0.21\textwidth}
    \includegraphics[width=\textwidth]{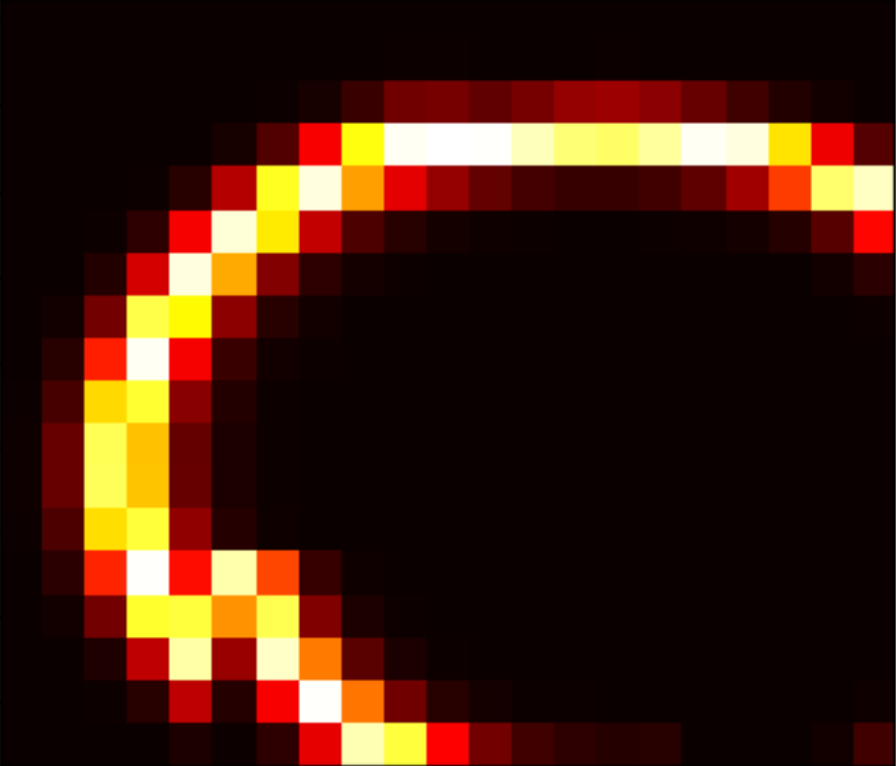}
    \caption{Feasible or Infeasible}
    \label{fig:sampling:two_class}
  \end{subfigure}
  \caption{Typical sampling patterns for uncertainty sampling
  techniques. \textbf{(a)} Pattern common to Least Confident, Minimum Margin,
  and Maximum Entropy when considering each ProMP in the ProMP library to be its
  own class, as described in
  Section~\ref{sec:experiment_results:qualitative}. \textbf{(b)} Pattern when
  considering only two classes, Feasible (ProMPs) and Infeasible (GMM), as
  described in Section~\ref{sec:experiment_results:regions}. Lighter values are
  more likely to be sampled.}
  \label{fig:sampling}
\end{figure}

\begin{figure*}
  \includegraphics[width=\textwidth]{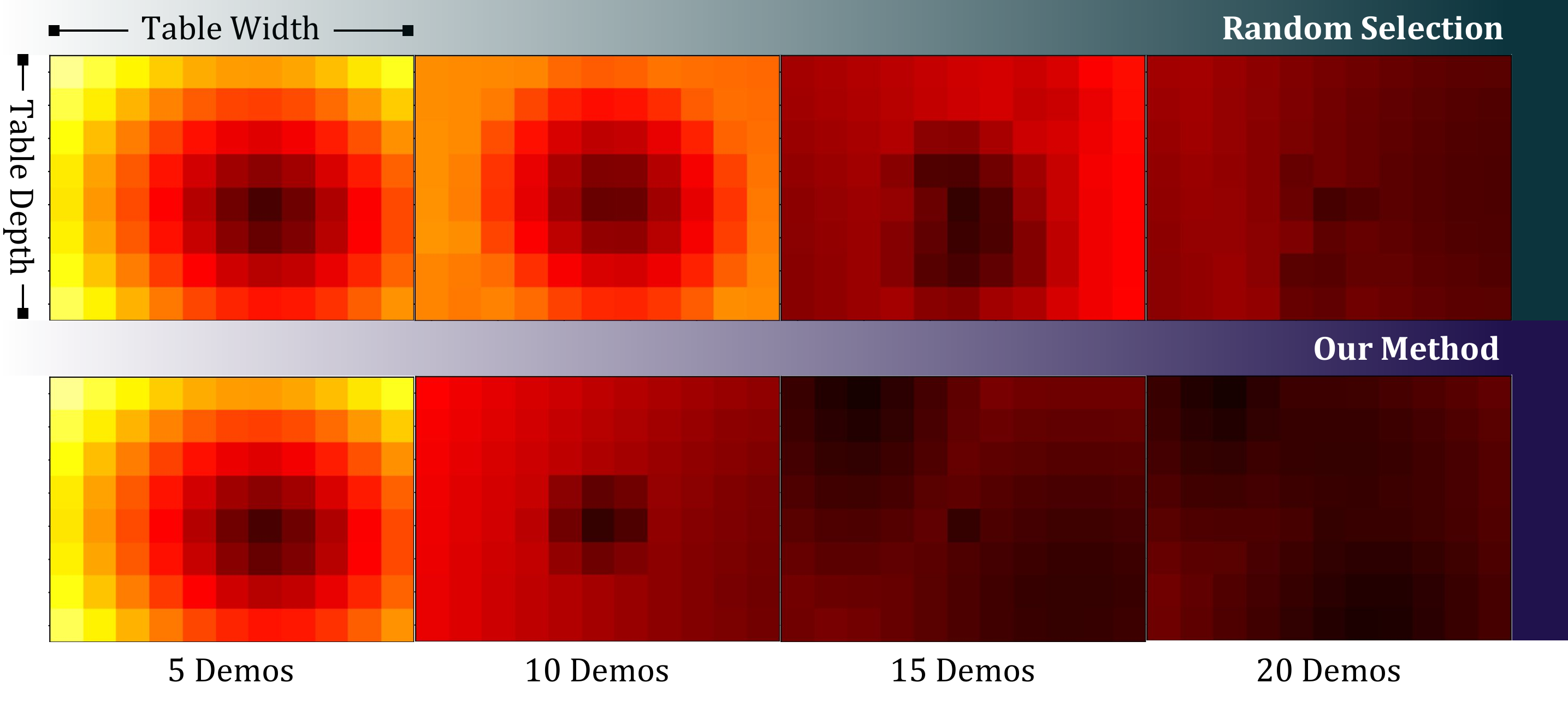}
  \caption{Sequence of heatmaps over a top-down view of the table surface
  illustrating the rate at which Greatest Mahalanobis Distance drops while
  performing random selection (top row) versus performing our optimization
  (bottom row). Lighter colors indicate higher Greatest Mahalanobis Distance
  (more likely to be sampled). Images show increasing number of acquired
  demonstrations going from left to right.}
  \label{fig:mahal_random}
\end{figure*}

\section{Experimental Results}
\label{sec:experiment_results}

\subsection{Qualitative Comparison of Uncertainty Sampling}
\label{sec:experiment_results:qualitative} 

We perform a qualitative comparison of the four uncertainty sampling methods
described in Section~\ref{sec:methods:active_learning}. We analyze the
progression of the uncertainty sampling metrics over the grid data space
described in Section~\ref{sec:experiment_setup} as more demonstrations are
achieved. As seen in Figure~\ref{fig:sampling:multi_class}, Least Confident,
Minimum Margin, and Maximum Entropy each tend to fixate selection on the
boundaries between ProMPs. Once at least two neighboring ProMPs become
well-estimated enough to produce meaningful probability measures, they begin to
``compete'' over the territory covered in part by both ProMPs. This behavior is
not desirable for the purpose of promoting task generalization over the entire
space. As such, we found that while these measures are the go-to objective
functions for uncertainty-sampling approaches in supervised active
learning~\cite{settles2012active}, they do not provide a suitable mechanism for
guiding the creation of a ProMP library that can generalize well over a given
space.

We propose the Greatest Mahalanobis Distance, described in
Section~\ref{sec:methods:active_learning}, as an alternative to these standard
measures. As seen in Figure \ref{fig:mahal_random}, the Mahalanobis distance
objective tends to converge to low values instead of becoming heightened on
boundaries between ProMPs. Even if a task instance can be achieved by multiple
ProMPs (i.e. the instance exists near a boundary between two ProMPs), its
minimum Mahalanobis distance is unaffected by such competition. We submit that
this behavior makes Greatest Mahalanobis Distance the most suitable measure
among the four compared for active learning of ProMPs, as it will tend to drive
the learning into regions that have not been explored, instead of fixating on
boundaries between already well-estimated regions of the task space.

\subsection{Task Success on Execution}
\label{sec:experiment_results:execution}

In order to demonstrate the efficacy of the Greatest Mahalanobis Distance
measure for active learning, we compare our method against randomly selecting
task instances on task executions on the robot as described in
Section~\ref{sec:experiment_setup}.  In order to account for randomness in the
learning process, we perform ten trials of learning ProMP libraries over the
space. We then chose the ProMP library that achieved the median performance on a
validation metric for testing on the robot.

We use the recorded demonstrations over the discretized space to perform the ten
learning trials. In each trial we use a random seed for the random sample
generation, and we use the same seed to generate a small set of initial samples
to initialize our active learning method. For each trial, we generated task
instances and collected 25 demonstrations for each method. We then ranked the
capability of the ProMP libraries by the value of the Greatest Mahalanobis
Distance computed over all task instances for use as our validation metric.

We generated a test set of ten random planar object poses to attempt with each
comparison method. We emphasize that the test poses were generated from a
continuous set, i.e. they are not selected from candidates in the discretized
space, and as such they are not likely to be identical to any instances the
methods received demonstrations for. The object was tracked and placed on the
table by the user to align with the coordinate frame of the generated instance,
as described in Section~\ref{sec:experiment_setup}. For each method, the most
likely ProMP and condition point to produce task success were selected based on
the Greatest Mahalanobis Distance measure for that object pose. The resulting
ProMP policy was then executed. We used task completion as our metric of
success, where the task is considered successfully completed if the object
remains in the robot's grasp after the lifting phase described in
Section~\ref{sec:experiment_setup} has completed.

Random selection of task instances resulted in only 2 out of 10 successful
grasps. 3 of the instances were attempted but quickly failed due to the robot
knocking the object off the table or pushing the object away as the fingers
started closing around it. The other 5 instances could not even be attempted due
to safety concerns in watching the execution previews in rviz. These were
primarily cases where the hand was clearly going to collide with the object or
table at a high velocity, risking potential damage to the robot hand. In
summary, random selection resulted in 20\% success, 30\% failure, and 50\%
infeasible due to safety concerns.

Our Greatest Mahalanobis Distance approach resulted in 6 successful grasps, 3
failed grasps, and only 1 infeasible instance due to safety concerns. Only 1 of
the failed grasps was due to the robot missing the grasp entirely due to pushing
the object away when the fingers close. The other 2 failures were attempted
overhead grasps in which the robot reached a suitable pre-grasp and closed the
fingers around the upper portion of the drill, but then proceeded to drop the
object on the lifting phase. The infeasible instance was due to what was a clear
collision between the fingers and the object at a high velocity. To summarize,
our method resulted in 60\% task success, 30\% failure, and only 10\% were
infeasible.

We note that the recorded demonstrations were generally either an overhead grasp
towards the head of the drill, or a side grasp radially located about the drill
handle. Overhead grasps were more suitable when the object was located closer to
the base of the robot, whereas side grasps were more appropriate the further the
object was located from the base. However, from the user's perspective, overhead
grasps were significantly more difficult to demonstrate successfully. This is
primarily due to the weight of the drill requiring a precise grasp pre-shape
from above to fully enclose the drill head without losing grip on the lifting
phase.

\subsection{Learning Feasible and Infeasible Task Regions}
\label{sec:experiment_results:regions}

In some situations the boundaries of the context region \(\mathcal{C}_d\) to
generalize over may not be explicitly known a priori. For such cases we propose
a minor extension to our approach enabling the robot to learn an explicit
infeasible region \(\mathcal{R}\) to avoid. We propose modeling this region
using a Gaussian mixture model defined on the context space \(\mathcal{C}\).

To formulate this as an active learning problem we treat the learned mixture of
ProMPs as a single positive class, with class probability defined by
Equation~\ref{eqn:promp_mixture}, and the GMM to represent the negative
class. When asked to provide a sample the user provides a demonstration as
before if the sample represents a point in the feasible region; otherwise the
user simply labels the point infeasible and the active learner provides a new
sample. In two-class cases, the Least-Confident and Minimum-Margin uncertainty
sampling methods are equivalent to Maximum Entropy~\cite{settles2012active}.

Figure~\ref{fig:sampling:two_class} visualizes the maximum entropy associated
with a feasible-infeasible learning trial. In the case pictured an obstacles
sits in the center of the table, which the robot should not collide with. We see
that the points of highest entropy (lighter colors) lie near the boundary
between this infeasible center region and the surrounding areas, known to be
feasible from example demonstrations. Thus the maximum entropy metric proves
useful in this scenario, selecting samples to refine the boundary between the
neighboring feasible and infeasible regions.


\section{Conclusion}
\label{sec:conclusion}

We have presented a framework for active learning of a library of Probabilistic
Movement Primitives from demonstration. Our method leverages existing active
learning techniques while utilizing the information encoded in the ProMPs to
compute the active learning measure guiding sample selection. We demonstrated
with real-robot experiments that our method provides an advantage over randomly
choosing demonstrations over the space in which generalization is desired. Our
method provides an uncertainty estimate of task success over a given region,
enabling the robot to be deployed to situations where a teacher may not be
available, e.g. remote missions in space.

In this paper, we only considered task generalization over a static
environment. In future work, we will explore adapting our methods to dynamic
environments in which task constraints vary over time, such as obstacles that
are not fixed features of the environment. Additional future work could examine
incorporating a more informed prior for classifying feasible and infeasible
regions that either leverages knowledge of an environmental map or could be
learned and transferred from previous tasks.


\bibliographystyle{IEEEtran} 
\bibliography{references}

\end{document}